# Real Time Strategy Language Version 1

Roy Hayes
Peter Beling
William Scherer

## Introduction:

Over the last decade researchers have begun focusing on artificial intelligence (AI) for real time strategy (RTS) games. RTS are popular computer games, which feature developing and supporting an army of different unit types and buildings. Unlike classical games, such as chess, players perform their actions in real time as oppose to turn based decisions. Additionally, players move simultaneously controlling potentially hundreds of units and buildings. To increase the difficulty of these games a player's knowledge of the playing map is restricted to a small area around their military assets and those of their allies.

Researchers have generally limited their AI implementation to subcomponents of games, such as resource management or path finding [1]. That is to say, researchers have sought to develop a solution to a single aspect of RTS game. In recent years there has been a push to develop AI systems capable of playing the full game. However, these implementations are limited to playing a single RTS game. This leads to researchers leveraging expert knowledge and game specific information. Although, this results in improvement for the specific game it does not address the root concern of AI research, which is to develop agents that can autonomously interpret, learn, and apply new knowledge. In essence AI RTS agents are improving because of more efficient programing and better human implemented strategy and not because improvements in artificial intelligence.

The authors' assert the need for Real Time Strategy Language. Such a language would allow for the creation of an agent that can autonomously learn how to play a previously unknown game. These general gaming agents have been developed for relatively simple games such chess, tic-tac-toe, and more recently poker, they are known as General Game Playing (GGP) agents [2]. By allowing these GGP agents to play against one another over a variety of games it becomes apparent, which AI system is more adept at learning and applying knowledge.

There are several advantages to developing general RTS agents, which leverage the GGP methodology. As these agents improve, measured by win percentage over both human and other AI opponents, it will be due to their ability to gather and apply new knowledge. This will demonstrate an improvement in our collective AI knowledge. Additionally, it will allow for faster development of new RTS games. Developing the AI scripts found in today’s games is a resource intensive project. The speed of game development will increase as AI systems become capable of learning how to play new RTS games. Lastly, general RTS agents will improve player experience. Traditional game AI systems are non-adaptive. Therefore, once the player has determined a weakness it becomes trivial to win. General RTS agents implement an adaptive learning system, thereby mitigating this problem [3].

A Real Time Strategy Language needs to be developed first, which will allow for the development of a GGP agent for the RTS genera of games. This paper proposes a RTS language. The authors' appreciate the complexity of RTS games and therefore future iterations will expand on the language. It should be noted that this language is not designed encompass Non-Real Time Strategy Games.

## Language Structure

The game description language was originally designed for turn based perfect information, symmetric games. It was later augmented to include imperfect information game. Each turn a player is given propositions, which are true in the given state. Additionally, the player is given possible steps that can be made from that state and how they will affect the game. The problem with this formulation is that as the game increase in complexity so does the number of propositions needed to describe the game state. In addition, the number of possible actions and transitions also increase with complexity [5].

Real time strategy (RTS) games are not conducive to the original format of GDL. Turn based updates are unfeasible for RTS game because they are played over a continuous timeframe. Furthermore, a player can control hundreds of buildings and units, leading to an exponential growth in possible actions and state transitions. Therefore, defining propositions for an RTS game's possible states and transitions is impracticable.

For an autonomous system to be competitive it must be able to act based on an internal view of the world. Parsing updates takes time, which limits the autonomous agent's ability to perform useful actions in the RTS game. Lewis et al, found a correlation between winning a game and the action rate of players [6]. A player that performs more actions than their opponent improves their position faster and thus giving them an advantage. Although, this is correlation and not causation it highlights the need for efficient updates, which is not possible under the current GDL language format.

The autonomous system could reconcile any errors in its internal view with periodic updates. Such a system would need a basic understanding of the game, prior to learning game specific information. In other words, the system should understand it is playing an RTS game and can assume that an army is made up of units, which will occupy a world. This allows for time saving domain knowledge to be added into the system.

As aforementioned, currently there is no autonomous agent that can beat an expert human in RTS games. Therefore, domain knowledge should be available to the autonomous agent. Leveraging domain knowledge limits the amount of information the autonomous agent needs to learn and may result in better performance. Additionally, it does not violate the core goal of AI research, which is to develop agents that learns and utilizes knowledge autonomously. An agent will still need to learn and utilize game specific information. To develop such an agent a RTS language needs to be created. The next section proposes a Real Time Strategy language.

## Faction Description

The game description can be split broadly into two segments. The first segment described here is the faction description. This describes the factions that can comprise the game, along with their buildings and units. As the description below demonstrates, Factions contain Human and Orc. Meaning there consist two factions named Human and Orc respectively. It is important to note that the Human and Orc is game specific information and can easily be renamed. The Resources denote what resources are available in the game. At the beginning of the game a player will have 100 pieces of wood for use. As with factions resources are game specific information

```
<Factions>
	Human
	Orc
```

```
</Factions >
<Resource>
        <Wood> 100 </Wood>
        <Gold> 100 </Gold>
        <Oil> 10 </Oil>
        <Food> 5 </Food>
</Resource>
```

In many RTS games races have different buildings and units, with different abilities. To accommodate game specific tags are used specify the properties of that object. Below is an example of a building specification.

```
<Humans>
        <Building>
                <Town Hall>
                        <UniqueID>TownHall1</UniqueID>
                        <Health Point>1200</Health Point>
                        <Terrain>
                                Ground
                        </Terrain>
                        <Action>Idle</Action>
                        <Shape>
                                <Square> 2 </Square>
                        </Shape>
                        <Position>
                                <X,Y>120,120 </X,Y>
                        </Position>
                        <Vision> 1 </Vision>
                        <Build Speed> 30 </ Build Speed>
                        <Enemy></Enemy>
                        <Require>
                                <Resource>
                                        <Wood> 800 </Wood>
                                        <Gold> 1200 </Gold>
                                </Resource>
                        </Require>
                        <Upgrade>Keep</Upgrade>
                        <Purpose>
                                <Process>
                                        <Resource>
                                                        Wood
                                                        Gold
                                        </Resource>
                                </Process>
                                < Build > Peasants </ Build >
                        </Purpose>
```

```
		</Town Hall>
	</Building>
</Human>
```

The building described is a town hall, which can be constructed by the faction known as human. The town all is comprised of 1200 health points, is constructed in the shape of a square, with each side being 2 units in distance. Enemies passing within in 1 unit of distance will be seen. It will take 30 seconds for the building to be constructed and will require 800 gold pieces and 1200 wood pieces. This structure can be upgraded to a Keep. The town hall has several purposes; it processes wood and gold adding them to the user bank. Additionally, it can train the unit know as Peasants.

The above description contains a mixture of keywords and game specific information. The tags that are bolded are keywords and can be found in the definition section. As stated earlier inner tags are properties of outer tags. That is to say Humans can construct buildings and one of their buildings is known as a Town hall, which requires 800 pieces of wood to construct. This hierarchal structure allows buildings to be described using a combination of keywords and game specific information. The same can be done for units.

```
<Human>
	<Unit>
		<Elvin Archer>
			<Health Point> 40</Health Point>
			< Build Time> 15 </ Build Time>
			<UniqueID> Archer1</UniqueID>
			<Armor>
				<Shield> 4 </Shield>
			</Armor>
			<Shape> Circle </Shape>
			<Size> 0.5 </Size>
			<Enemy></Enemy>
			<Action>Idle</Action>
			<Position>
				<X,Y>120,120 </X,Y>
			</Position>
			<Attack>
				<Arrow>
					<Range>4</Range>
					<Damage>3-9 </Damage>
					<Recharge>2   </Recharge>
				</Arrow>
			</Attack>
			<Terrain>
				Ground
			<Terrain>
			<Vision>5</Vision>
			<Speed>3</Speed>
```

```
                <Require>
                        <Resource>
                                <Gold> 500 </Gold>
                                <Wood> 50 </Wood>
                                <Food> 1 </Food>
                        </Resource>
                </Require>
        </Elvin Archer>
    </Unit>
</Human>
```

Above is a description of the Elvin Archer. The description provides information of the Elvin Archer Defensive and offensive capabilities. For example, the archer carriers a shield which will decrease any attack damage by 4 points. Additionally, the archer takes two seconds between firing arrows. These arrows can be targeted at units in the air and on the snow. The previous two examples demonstrate how units and buildings can be described in RTS games. More advance descriptions will be covered in the keyword section. The next section will discuss how to describe the environment.

## Environment Description

Maps describe the environment in which the game is played. This is where different terrains are specified, as well as the amount and location of different resources. Below is a brief example of the map specification system.

```
<Map>
    <Name> Hills </Name>
    <(0,0)>
        <Terrain>
            Ground
            High
            Air
        </Terrain>
        <Gold>1000</Gold>
    </(0,0)>
    <(0,1)>
        <Terrain>
            <Wood>300</Wood>/Ground
            Low
            Air
        </Terrain>
    </(0,1)>
</Map>
```

The above specification demonstrates several complex features. In the first cell the terrain is specified as high ground and air but there is a gold deposit worth 1000 units on it. That is to say, units that can pass through air or ground can occupy this space. In the second cell the terrain starts as wood but once the 300 units of wood is removed it becomes low ground, this allows for the generation of temporary obstructions.

When an RTS AI receives an update from the game engine about the state of the environment, there is a large amount of data that is given. However, the XML structure allows for rapid parsing of the data into relevant information. For example, in a standard 128 X 128 map there is information about 16,384 grid cells. However, like human players a RTS AI agent will only care about a small subset of these cells at any one time, specifically the cells that are within the vision of its buildings and units.

## Game Management

The RTS game management system has already been developed for artificial intelligence competition. The game management system accepts the connection of an AI system. The system designates which faction it intends to play as. The game manager selects a map, informs the AI system of their opponent's race, and sends a signal when the match begins.

The game management system maintains the true state of the game. That is to say the game management system keeps track of units, buildings, resources and map conditions. AI systems store separately what they believe to be the true state of the world. By querying the game management system for updates about the current state of the map, a player's stored state can be brought into alignment with current state of the game. Below is a diagram of the system combined with the RTS language. The aforementioned system has been used by AI StarCraft competition for the past several years.

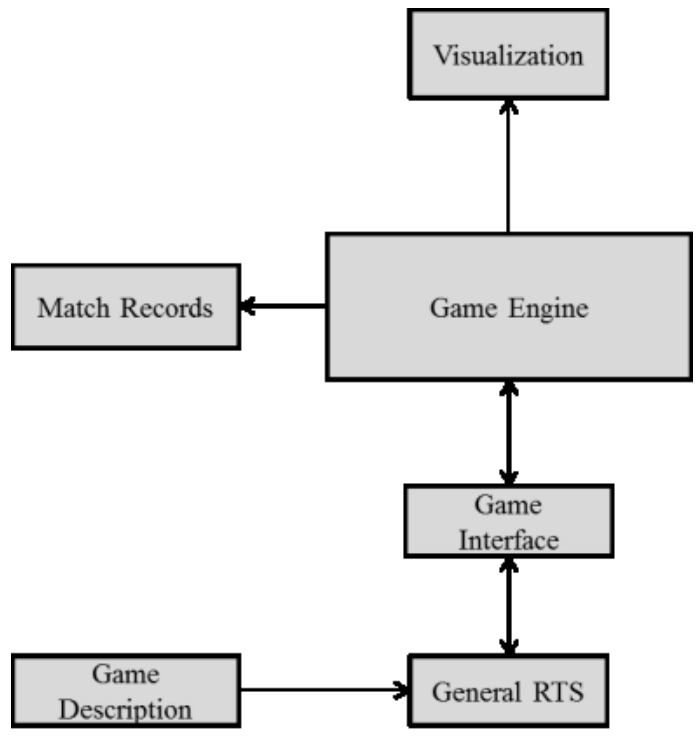

The authors' contend that there is no need for the underlying system to be modified. Rather the messages should be standardized, thus allowing an AI to interact with any RTS game. For standardization purposes updates of the system should be given in the aforementioned formats. The information given to the AI system should be limited by the vision of the AI's current units and buildings.

In RTS games there are several actions that a player can take. The RTS language has developed keywords for the following task:

1. Construct Building – Construct(buildingName, X-position, Y-position)
2. Move a Unit or Building – Move(Unique ID, X-Position, Y- Position)
3. Train – Train(Training location Unique ID, Name of unit/upgrade),
4. Gather Resource – Gather(Unit Unique ID, X-Position, Y-Position)
5. Attack – (Allied Unique ID, Enemy Unique ID)
6. Action – (Game Specific Action, Allied Unique ID [], Enemy Unique ID [], X-Position [], Y-Position []) the [] denotes an array
7. Update

The next section will give definition of keywords in the language.

## Keyword Definition

**#, #**

This specifies the grid environment. It is assumed that x and y are real numbers and the origin is the top left corner. Additionally, the property of the grid box that is formed by 4 pairs of x and y coordinates is defined by the lowest x and y coordinate.

```
<0, 0 >
        <Terrain>
                Sea
        </Terrain>
</0, 0>
<1, 0 >
        <Terrain>
                Grass
        </Terrain>
</1, 0>
<0, 1 >
        <Terrain>
                Grass
        </Terrain>
</0, 1>
<1, 1 >
        <Terrain>
                Dirt
        </Terrain>
</1, 1>
```

| | |
|---|---|
| 0,0 | 0,1 |
| 1,0 | 1,1 |

---

**Action**

Action states the current action being taken by a building or unit. The list of possible actions a unit or building can do is given below.

<Moving> x,y </Moving> - position the unit or building is moving to
<Attacking> UniqueID </Attacking> - building or unit that selected entity is attacking
<Gathering> Resource </Gathering> - Resources the unit is gathering. This includes the process of bring the resource back to the processing site
<Build>Building1<Build> - Tells what the unit is constructing, training, or researching.
Game Specific Action – This is defined as a unit or building game specific ability
Idle

---

**Armor**

Armor detracts from the damage done from an attack. Armor can be specified as a real number or a percentage. Armor can be universally specified or made attack specific. The example below armor lowers the attack damage of any attack by 2, except for Arrows and Swords. The attack damage of Arrow damage is decreased by 3% and sword damage is lowered by 5.

```
<Armor>
	2
	<Arrow> 3% </Arrow>
	<Sword> 5 </Sword>
</Armor>
```

---

**Attack**

Attack encompasses all the attributes of an attack. It defines the range, shape, size, damage, recharge rate, attackable terrain and requirements for an attack. Multiple attacks can be specified for each entity. Below, the attack requires 5 mana. Mana is game specific information and can be specified for each unit or the faction as a whole.

```
<Attack>
	<Attack 1>
		<Range>4</Range>
		<Damage>3-9 </Damage>
		< Recharge >2</Recharge>
		<Shape> Point </Shape>
		<Terrain>
			Terrain 1
		</Terrain>
	<Require>
		<Mana> 5 </Mana>
	</Attack 1>
</Attack>
```

---

**Build**

Build list the possible things a unit or building can construct, train, or research. Possible buildable items are separated by a newline character.

```
<Build>
	Tank
	Jet
</Build>
```

---

**Building**

The Building keyword encompasses all the information about buildings for a given faction.

---

**Building Time**

Building time is the time it takes in seconds to construct, train, or research something. Building time is assumed to be a real number.

---

**Contain**

Contain is a tag that is used for buildings or units that can contain other units, such as a transport ship. This tag allows you to specify restrictions on which units can enter the structure. A total of 8 weight units are available for this structure. The combine unit weight of any units inside the structure cannot exceed 8. Additionally, only units with light armor can enter the structure.

```
<Contain>
	<Weight> 8 </Weight>
	<Armor> Light </Armor>
</Contain>
```

---

**Damage**

Damage is the amount of damage caused by an attack. Damage values are given with a minimum damage and a maximum damage. Damage can also be specified universally or for individual units. If specified for individual unit, the universal damage is overridden for that unit. It is assumed the maximum damage is always caused unless armor negates the damage.

```
<Damage>
	2-5
	<Horse> 4 -9</Horse>
</Damage>
```

---

**Distance**

Distance specifies when something must occur within a given distance or outside a given distance. An action must occur a distance of 1 unit away from the structure but no more than 5.

```
<Distance>
	<Less> 5 </Less>
	<Greater> 1 </Greater>
<Distance>
```

---

**Enemy**

Enemy list the current enemies in an entity's field of vision. The amount of information given in this tag is game specific. The health point information is optional. The game designer can choose what information to give.

```
<Enemy>
	<Horse>
		< UniqueID>
			<Health Point> 100 </Health Point>
		</UniqueID>
	</Horse>
</Enemy>
```

---

**Faction**

Factions specify the possible playable groups. In most RTS games there are multiple factions that can be played each with their own buildings, units, and effective strategy. Each Faction is separated by a newline character.

```
<Faction>
```

Faction 1
Faction 2
</Faction>

---

**Gather**

Gather specifies what resources the unit is capable of gathering. The first number is the current amount of resources the unit is carrying; the second number is the maximum amount the unit can carry.

```
<Gather>
	<Resource 1> 50-100 </Resource 1>
</Gather>
```

---

**Health Point**

Health point describes how much life an object has remaining. It is assumed that health point is a real number. Once the health point of an object falls below 0 it is removed from the game.

```
<Health Point>1200</Health Point>
```

---

**Limit**

Specifies the number of times an ability can be used. The value must be an integer.

```
<Mine>
	<Limit> 4 </Limit>
</Mine>
```

---

**Map**

Map indicates that anything within this tag is considered to be related to the environment being played on. The RTS Language assumes that the game is being played on a 2 dimensional board. 3 dimensions can be added by specifying multiple terrains for a single coordinate space.

```
<Map>
	<Name> Name of Map </Name>
	<#, # >
		<Terrain>
			Ground
			Air
		</Terrain>
	</#, # >
</Map>
```

---

**Modify**

Modify tells the computer that performing this action will modify a given property. In some games, the attacking an opponent on high ground decreases your attack strength by some percentage. This can be encoded using the hierarchal structure. The example below shows that units occupying the low ground will have their attack damage decreased by 25% when attacking units that occupy high ground. The same can be done for movement speed, vision, etc.

```
<Terrain>
	<Low >
```

```
        <Modify>
            <Attack>
                <High>
                    <Damage>-25%</Damage>
                </High>
            </Attack>
        </Modify>
    </Low >
</Terrain>
```

---

**Movement**
This tag is optional, it specifies movement values. If this tag is not present, it is assumed units occupy and move over the same terrain at a given speed. However, in some games units or buildings may land on the ground and move through the air. This is demonstrated in the example below.

```
<Terrain>Ground</Terrain>
<Movement>
    <Terrain>Air</Terrain>
<Movement>
```

---

**Name**
Name is the name given to the current map. It is assumed that the AI system will be trained to play on each specific map. At the beginning of each match the name of the map will be given to the AI system, allowing it to play the strategy for that map. See Map for example.

---

**Prepare**
Prepare indicates that a building is required to be placed on a resource to extract it. This is analogous to an oil platform being placed over an oil well. This facility prepares a resource for extraction it may or may not process the resource. Resources separated by newline character.

```
<Prepare>
    Resource 1
</Prepare>
```

---

**Process**
Process indicates that a building is a warehouse for a specific resource. Resources delivered to this building will be added to the user’s total. See purpose for example.

---

**Purpose**
Purpose tag will give information about a buildings function. This includes all possible units that can be trained in the building. Additionally, it will designate if a building prepares or processes resources. Lastly, it will tell all possible research that can be found studied at the building.

```
<Purpose>
    <Process>
        <Resource>
            Resource 1
        </Resource>
    </Process>
```

<Build>
Unit 1
Tech 1
</ Build >
</Purpose>

---

**Range**
Range defines the maximum distance a unit or building can be and still be attacked or repaired, assumed to be a real number.
<Range> 1</Range>

---

**Recharge**
Recharge is the time between attacks in seconds. This can be thought of as the time needed for an archer to draw and fire another bow.
<Recharge>3</Recharge>

---

**Repair**
Repair is the amount of health points a unit can restore to another unit/building every second. Like damage this can be defined universally or made unit specific. If a unit is specified it overrides the universal repair rate and range for only that unit.

```
<Repair>
	2
	<Range>1</Range>
	<Horse>
		1
		<Range>2</Range>
	</Horse>
</Repair>
```

---

**Require**
Require specifies what is required to train, build, or research something. This can include having the required amount of resources available for use, having necessary buildings constructed, or completion of technology research. Notice because Tech 1 is game specific information it does not have to go into a keyword tag.

```
<Require>
	<Resource>
		<Resource 1> 500 </ Resource 1>
		< Resource 2> 50 </ Resource 2>
	</Resource>
	<Building>
		Building 1
		Building 2
	</Building>
	Tech 1
	Tech 2
</Require>
```

**Resource**

Resources are any items that must be collected from the map to construct units or buildings. This can also be extended to encompass items that are generated from the construction of buildings and are required for constructing other units or buildings, such as food from farms. Each Resource is separated by a newline character. Additionally, the value with in the game specific resource tag indicates the current amount of that resource available to the player.

```
<Resource>
        <Resource 1> 50 </Resource 1>
        <Resource 2> 300 </Resource 2>
        <Resource 3> 25 </Resource 3>
</Resource>
```

**Shape**

Entities and attacks possess a shape. There are multiple shapes that can be used as keywords. The x,y position is always assumed to be the center of the shape.

*Point* – Point is a single point in space.

*Square* – The size of the square specifies the length of the square side.

```
<Square>
        <Size> 5 </Size>
</Square>
```

*Rectangle* – The size of a rectangle is defined by two numbers. The first number defines the sides that run parallel to the attacking entity, the second number define the perpendicular side.

```
<Rectangle>
        <Size> 5-10</Size>
</Rectangle>
```

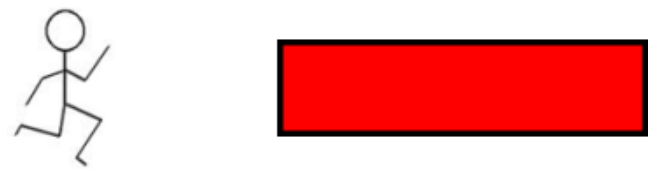

*Circle* – The size of a circle defines the radius of the circle.

```
<Circle>
        <Size> 5 </Size>
</Circle>
```

*F_Cone* – Defines a cone with the base at the far end of the cone. Two numbers define the cone, the first number defines the height and the second number defines the length of the base.

```
<F_Cone>
        <Size> 10-5</Size>
</F_Cone>
```

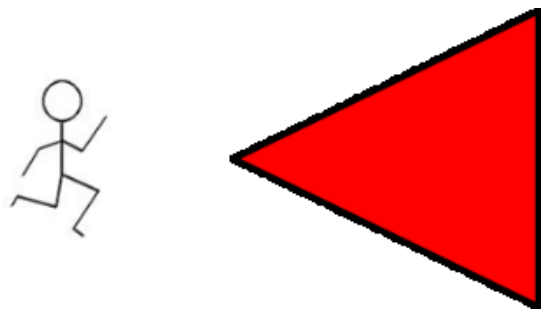

*B_Cone* – Defines a cone that is an F_Cone flipped 180 degrees.

```
<B_Cone>
        <Size> 10-5</Size>
</B_Cone>
```

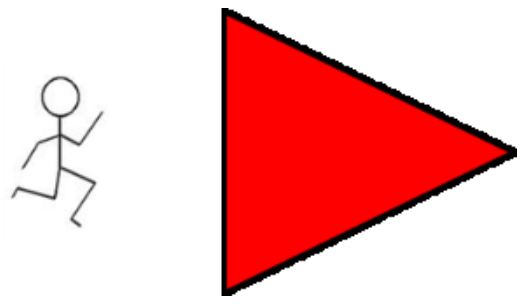

---

**Size**
Size specifies the size of the shape. Depending on the shape there may be one or two numbers specified. Look at Shape for example.

---

**Speed**
Speed is the distance an entity can move per second. It is assumed the value is a real number. This may be slowed down due to obstructions, such as walls.

```
<Speed> 2 </Speed>
```

---

**Terrain**
The meaning of terrain changes depending on the tag it is encompassed by. For Maps it specifies the terrain that the current grid square is made up of. If it is encompassed by the Buildings tag it specifies the terrain the buildings can be built on and for units it specifies the terrain the units can move through and attack. Additionally, terrain can be specified to change depending on conditions. The example below shows a resource of 300 wood pieces. At the beginning of the game wood is the terrain but once the wood is removed snow becomes the terrain.

```
<Terrain>
        <Wood>300</Wood>/Snow
        Air
</Terrain>
```

---

**Time Limit**
This specifies the time any ability can stay active. It is given in seconds. See Game Specific Ability for an example.

---

**Weight**
This is an optional tag. It is used to specify the weight of a unit. The combine weight of all units within a structure cannot exceed a defined maximum. See contain for example.

---

**Vision**
Vision is the distance that an entity can see away from it, assumed to be a real number. Any enemies that cross within its field of vision will be added to its enemy tag for the duration the enemy is in the field of vision. Additionally, if the user asks for an update, map coordinates within its field of vision will be included in the update.

<Vision> 5 </Vision>

---

**Unit**
The Unit keyword encompasses all the information about units for a given faction.

---

**UniqueID**
Every building and unit is given a unique identification so the AI and the game engine can identify the correct unit.

---

**Upgrade**
Upgrade specifies the possible things the unit or building can be upgraded to.
<Upgrade>Keep</Upgrade>

---

**Game Specific Ability**
To define a game specific ability create a tag with the ability name inside the unit or building that will use it. Within the ability tag specify what is being altered and if there is any requirement for the ability. See example below. The example below demonstrates a special ability. A non-biological enemy can be given a speed of 0 and a weapon recharge of essentially infinite. This prevents the unit from moving or firing their weapon. This affect last for 12 seconds.

```
<Lockdown>
        <Enemy>
                <Recharge> 100000 </Recharge>
                <Speed> - 100% </Speed>
        </Enemy>
        <Require>
                <Enemy>
                        </Biological> False </Biological>
                </Enemy>
        </Require>
        <Time Limt> 12 </Time Limt>
</Lockdown>
```